# A Vertical Federated Learning Framework for Horizontally Partitioned Labels

Wensheng Xia
Peking University
xia.wensheng@pku.edu.cn

Ying Li*
Peking University
li.ying@pku.edu.cn

Lan Zhang
Michigan Technological University
lanzhang@mtu.edu

Zhonghai Wu
Peking University
wuzh@pku.edu.cn

Xiaoyong Yuan
Michigan Technological University
xyyuan@mtu.edu

## ABSTRACT

Vertical federated learning is a collaborative machine learning framework to train deep leaning models on vertically partitioned data with privacy-preservation. It attracts much attention both from academia and industry. Unfortunately, applying most existing vertical federated learning methods in real-world applications still faces two daunting challenges. First, most existing vertical federated learning methods have a strong assumption that at least one party holds the complete set of labels of all data samples, while this assumption is not satisfied in many practical scenarios, where labels are horizontally partitioned and the parties only hold partial labels. Existing vertical federated learning methods can only utilize partial labels, which may lead to inadequate model update in end-to-end backpropagation. Second, computational and communication resources vary in parties. Some parties with limited computational and communication resources will become the stragglers and slow down the convergence of training. Such straggler problem will be exaggerated in the scenarios of horizontally partitioned labels in vertical federated learning. To address these challenges, we propose a novel vertical federated learning framework named Cascade Vertical Federated Learning (CVFL) to fully utilize all horizontally partitioned labels to train neural networks with privacy-preservation. To mitigate the straggler problem, we design a novel optimization objective which can increase straggler's contribution to the trained models. We conduct a series of qualitative experiments to rigorously verify the effectiveness of CVFL. It is demonstrated that CVFL can achieve comparable performance (e.g., accuracy for classification tasks) with centralized training. The new optimization objective can further mitigate the straggler problem comparing with only using the asynchronous aggregation mechanism during training.

## 1 INTRODUCTION

Federated learning [1][2][4] enables multiple parties to collaboratively train a machine learning model with privacy-preserving mechanisms. Conventional federated learning (horizontal federated learning) assumes that each party has the same set of features but different data. Recently a more realistic scenario has been investigated in federated learning, where different features are vertically distributed across participants, i.e., vertical federated learning [10, 11, 12, 14, 15, 16, 20]. In vertical federated learning, different parties hold the same set of data, while each party only has a disjoint subset of features. Vertical federated learning is attracting more and more attention due to its great potential of promoting broader participation in many industrials where privacy is paramount, such as healthcare [25], finance [26], and banking [31]. Thus, in this paper, we focus on vertical federated learning.

Unfortunately, most existing vertical federated learning methods have not addressed two key challenges. First, most vertical federated learning methods assume that at least one party holds the complete set of labels of all data samples. However, this assumption often violates the nature of many practical scenarios, where the parties usually have only partial labels. Similarly to the raw features, the partitioned labels cannot be shared among parties due to privacy concerns. For example, several hospitals, clinics, and laboratories plan to collaboratively build a machine learning model without compromising patients' private data. A vertical federated learning model is trained on the features collected from different parties (e.g., patients' regular visits in clinics, genomic sequencing in laboratories, and radiology images in the hospitals). Each party does not own the complete set of features. Moreover, the medical diagnosis (i.e., labels) are made by different hospitals and cannot be shared due to privacy regulations. Therefore, none of the hospitals owns all patients' diagnosis (i.e., the complete set of labels). Under this practical scenario, it becomes non-trivial to adopt the existing vertical federated learning methods. Vertical federated learning requires labels to update model parameters via end-to-end backpropagation. However, the partial labels held by each party greatly affect the effectiveness of the vertical federated learning. Therefore, it comes the crux of designing a novel vertical federated learning algorithm, that can make good use of the partitioned labels across different parties to train models which has better performance than the models trained by only using partial labels.

Second, recent research has shown that the efficiency of federated learning can be substantially reduced due to the inevitable straggler problem. Stragglers usually have limited computational and communication resources, and hence, cannot participate in the training process timely. For example, compared with hospitals, some private clinics are usually equipped with less powerful resources and require more time to train the models, significantly slowing down the federated learning process. Compared with horizontal federated learning, such straggler problem becomes even more serious in the vertical federated learning setting, since vertical federated learning concatenates all the local models and strongly depends on the local features. Such straggler problem can be further exaggerated in the scenarios of partial labels. Although several works [10, 11, 14, 30] have investigated asynchronous aggregation mechanism in vertical federated learning, asynchronous aggregation is still insufficient in mitigating the straggler problem. First, although asynchronous aggregation reduces waiting time, the numbers of training rounds of the stragglers are still far less than the other parties that are rich in computational and communication resources, which will reduce the convergence rate of training. Second and more importantly, skewed distribution of labels is a natural phenomenon in reality. That is to say, the parties that own labels usually have different distributions of labels. Thus, when computational resources vary greatly among parties that hold labels, the trained model will no doubt be affected more by parties that own rich resources since these parties update models more times than the straggler parties during training. Therefore, the trained model cannot well capture the distributions of labels from stragglers, which affects effective participation of stragglers.

In this paper, we propose a Cascade Vertical Federated Learning (CVFL) framework to address the aforementioned challenges. Specifically, the model parameters on disjoint features are aggregated and updated via cascade training of two types of subnetworks in a bottom-up fashion. The bottom subnetworks are located in all parties in order to extract embedding vectors from features. The top subnetworks are located in parties that own the subset of labels and conduct the prediction. The cascade training addresses the necessity of labels for model updates in conventional end-to-end backpropagation. We propose a novel optimization objective to further mitigate the straggler problem in vertical federated learning leveraging the recent research in fairness-aware federated learning [21]. Our experimental results demonstrate that our proposed algorithm can train neural network models with minimum performance loss and mitigate the straggler problem.

Our contributions are summarized as follows.

- We design a cascade vertical federated learning framework, which enables multiple parties to collaboratively train a neural network model with partitioned features and labels. To the best of our knowledge, we are the first to develop vertical federated learning framework in the scenario where labels are partitioned across different parties.
- We propose a novel optimization objective which can accelerate the convergence of model training and improve the trained model performance to mitigate the straggler problem.
- We conducted extensive experiments on three widely-used datasets and the results showed the effectiveness of our proposed algorithm.

The reminder of this paper is organized as follows. In Section 2, we discuss related work. Then in Section 3, we formulate the problem. Section 4 provides a detailed explanation of our approach. Section 5 elaborates the details of experiments and results. Finally, we conclude this paper and discuss the future work in Section 6.

## 2 RELATED WORK

Privacy concern is one of the main reasons that prevent pervasive participation in collaborative machine learning. Federated learning, a new learning paradigm, has been proposed recently for collaborative machine learning without sharing private data [1], [2]. Federated learning allows multiple parties to train local models on their private data and share the model parameters to a server, and the server aggregates these parameters to obtain a global model. Hence, the privacy of each party's local data is preserved by local training. Most existing federated learning methods [3]–[6] [27] focus on the horizontal federated learning setting, where a large dataset are partitioned among different parties that share the same feature space.

Recently vertical federated learning has attracted much attention due to its pervasive use in real-world applications. In vertical federated learning, each party holds a disjoint subset of features to train a collaborative machine learning model. Most vertical federated learning usually shares intermediate computational results among each party and updates the model parameters using distributed stochastic gradient descent (DSGD) methods. Compared with horizontal federated learning, vertical federated learning is usually limited in communication efficiency and privacy preservation, since the global model is the concatenation of local models and the update of the model strongly depends on the participation of all parties. Early research on vertical federated learning assumes that all the parties own the complete set of labels of data samples [7]–[10] [20] so that all the parties can calculate the loss function and update the model locally. Recent work has relaxed this assumption and assumes that only partial parties or a trusted third party (e.g., a global server) owns the complete set of labels. For example, Chen et al. [11] proposed an asynchronous federated learning method to extracted a perturbed local embedding the global model. A server, which has the complete set of labels, can calculate all the gradients for the parties. However, the private information of labels can be exposed when the server is semi-honest (honest-but-curious). Cheng et al. [12] provided a secure mechanism for tree boosting algorithms (e.g., XGBoost [22]), when only one party owns the labels. They allow all the parties to derive locally optimal split and require the passive parties, who do not have labels, to transfer the split information to the active parties using additive homomorphic encryption [13]. Zhang et al. [14] designed a bilevel asynchronous parallel architecture to allow only one or partial parties to hold the labels (active parties). The parties, who don't have the labels (passive parties), transmit their intermediate results to the active parties via efficient tree-structured secure aggregation [7], [15].

In our paper, we investigate a more common yet challenging scenario where none of the parties (or the centralized server) own the complete set of labels. The partitioned labels across different parties lead to the daunting challenges in the model training and collaboration, since none of the parties can have the complete information to update the global model. The vertical federated learning becomes even more challenging in such scenarios when considering the straggler problem. The training performance may be greatly impacted by the stragglers who have partial labels but limited computational resources. In this paper, we address the straggler problem using a novel objective function in the local optimization.

Another line of federated learning methods focuses on the partially additive homomorphic encryption (HE) [13][20][23][28] and secure multi-party computation (MPC) [16][17][29]. Unfortunately, these methods can only apply to linear models. Taylor series approximation is required for the non-linear models, which inevitably degrades the models' prediction performance [18]. Moreover, such cryptosystems introduce additional computational cost and are extremely time consuming for resource-constrained parties in many real-world applications. Therefore, in our paper, we only consider the approaches that share intermediate computational results in vertical federated learning. As proved by Liu et al. [19], when the dimension of features of the party $m$ is larger than 2, it is secure to share intermediate computational results (e.g., embedding vectors) among parties. Since in this case, there exists infinite solutions that can yield the same intermediate result and it is impossible for adversaries to infer the model parameters and the exact raw data from the intermediate result. Therefore, our approach can preserve the privacy of model parameters and the raw data.

## 3 PROBLEM FORMULATION

In this paper, we consider a set of $M$ parties: $\mathcal{M} := \{1, \dots, M\}$ that participate in model training. A dataset of $N$ samples, $\{x_n, y_n\}_{n=1}^N$, are partitioned across $M$ parties. Each party is associated with a unique set of features. For example, the $m$-th block features $x_{n,m} \in \mathbb{R}^{d_m}$ of the $n$-th sample $x_n = [x_{n,1}^T, \dots, x_{n,M}^T]^T$ are maintained by the $m$-th party.

**Definition 1. Active Party.** We define the active party as the party that holds not only data features but also labels of data samples. The active party is the dominator during training since machine learning requires labels to derive the loss function.

**Definition 2. Passive Party.** We define the passive party as the party that only provides extra features during training but without labels of data samples.

**Definition 3. Fast Active Party.** We define the fast active party as the active party that has rich computational and communication resources.

**Definition 4. Slow Active Party.** We define the slow active party as the active party that has limited computational resources or communication resources. Therefore, the slow active party updates models far less times than the fast active party.

Particularly, we suppose that the first $L$ parties are active parties, which means that the labels are partitioned among these $L$ parties. For instance, the labels owned by the $l$-th active party are $\{Y_{n,l}\}_{n \in \mathcal{G}_l}$, where $\mathcal{G}_l$ is the index set of labels maintained by $l$-th active party and $|\mathcal{G}_l| = N_l, l = 1, \dots, L$. These $L$ active parties can further be divided into fast active parties and slow active parties according to their computational capabilities.

In order to prevent the possible privacy leakage, raw features and labels are not allowed to share across parties. Instead, each party $m$ trains an embedding model $\theta_m$, which maps the local raw features $x_{n,m}$ to a compact embedding vector $h_{n,m} := h_m(\theta_m, x_{n,m})$, $h_m$ represents the embedding function. Ideally, all parties want to solve the following regularized empirical risk minimization problem:

$$\min_{w_0, \boldsymbol{\theta}} F(w_0, \boldsymbol{\theta}) = \min_{w_0, \boldsymbol{\theta}} \left[ \frac{1}{N} \sum_{n=1}^N \ell(w_0, h_{n,1}, \dots, h_{n,M}; y_n) + \lambda \sum_{m=1}^M r(\theta_m) \right], \quad (1)$$

where $w_0$ denotes the prediction model that takes the concatenation of all embedding vectors as input to make prediction; $\boldsymbol{\theta} := [\theta_1^T, \dots, \theta_M^T]$ is the concatenation of all embedding models, $\ell$ denotes the loss function and $\sum_{m=1}^M r(\theta_m)$ is the regularization term. Note that the prediction model $w_0$ cannot be directly learned by any single active party in the scenario where labels are partitioned among active parties. All $L$ active parties that hold labels will actively launch model updates and all $M$ parties including active parties and passive parties will passively launch model updates. The problem that we want to solve in this paper can be stated as:

**Given:** A vertically partitioned data features $\{x_{1,m}, \dots, x_{n,m}\}_{m=1}^M$ distributed on $M$ parties and the class labels $\{y_n\}_{n=1}^N$ are also partitioned among $L$ active parties.

**Learn:** A set of embedding models and a prediction model that are collaboratively learned by all parties with privacy-preservation.

The embedding models extract embeddings from different parties' features and the prediction model conduct the prediction.

**Goal:** When training is converged, we require that the trained embedding models and the prediction model which performance (e.g., accuracy for classification tasks) are comparable to the entire model trained under centralized training. What's more, our algorithm should be more efficient than only using asynchronous aggregation mechanism when there are stragglers during training.

**Notations.** In order to make notations easier to follow, we give a summary of key notations of this paper in Table 1.

| Notation | Description |
| --- | --- |
| $\mathcal{M}$ | The set of parties that participate in training. |
| $M$ | The number of parties that participate in training. |
| $N$ | The number of data sample. |
| $x_n$ | The $n$-th sample's features. |
| $x_{n,m}$ | The $m$-th block features of $n$-th sample. |
| $L$ | The number of active parties. |
| $\mathcal{G}_l$ | The index set of labels maintained by $l$-th active party. |
| $N_l$ | The number of labels owned by $l$-th active party. |
| $\theta_m$ | The model parameters of embedding model of $m$-th party. |
| $h_m$ | The embedding function of $m$-th party. |
| $h_{n,m}$ | The embedding vector of the $m$-th block features $x_{n,m}$. |
| $w_0$ | The model parameters of the global prediction model. |
| $w_0^l$ | The model parameters of the local prediction model of active party $l$. |
| $\boldsymbol{\theta}$ | The concatenation of parameters of all embedding models. |
| $F$ | The objective function of the overall problem. |
| $F_l$ | The objective function of the $l$-th active party. |
| $\ell$ | The loss function. |
| $r(\theta_m)$ | The regularization term of the embedding model. |

**Table 1. Summary of Key Notations in this paper**

## 4 PROPOSED APPROACH

In this section, we describe the details of CVFL framework and the novel optimization objective for slow active parties. The pseudocode implementation of CVFL is given in Algorithm 1, Algorithm 2 and Algorithm 3.

### 4.1 Cascade Training Mechanism

The end-to-end backpropagation for neural network training needs labels to derive the loss function. When each active party only has partial labels, it can only partially update the parameters of neural network based on its own labeled data. Therefore, the key of designing a proper vertical federated learning algorithm to train neural network in the scenarios of partitioned labels is to decompose the loss function in end-to-end backpropagation.

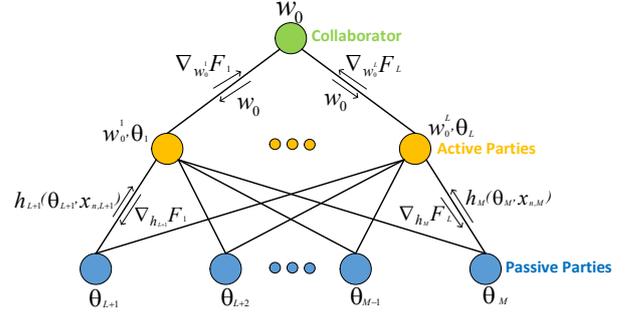

**Figure 1: Overview of CVFL framework. Each node represents a party and there are $L$ active parties and the other parties are passive parties. Each party has a bottom subnetwork and the active party also has a local top subnetwork. Each active party is connected with other parties so as to collect the embedding vectors from other parties and send gradients back. The collaborator is connected with all active parties so as to aggregate the gradients of all local top subnetworks.**

To address this problem, we design cascade training mechanism that breaks down the entire neural network into two types of subnetworks in a bottom-up fashion. As illustrated in Figure 1, there are three types of participants in CVFL, i.e., active parties, passive parties and the collaborator. As described before, each party $m$ trains a bottom subnetwork $\theta_m$ that learns to map the high-dimensional raw features of its own into low-dimensional compact embedding vectors. Each active party is also in charge of training a local top subnetwork $w_0^l$ that learns to predict the result by taking the concatenation of all embedding vectors as input. Then each active party uses its own labels to derive the loss and calculate the gradients with respect to its top subnetwork and other parties' bottom subnetworks. The collaborator receives and aggregates the gradients of all active parties' local top subnetworks $\nabla_{w_0^l} F_l, l = 1, \ldots, L$ to update the global top subnetwork $w_0$. Note that the collaborator does not need to be a trusted third party. Our design of CVFL framework can still preserve data privacy when the collaborator is semi-honest (honest-but-curious).

When an active party actively launches updates, it will send the gradients of its top subnetwork to the collaborator. After that, the collaborator will use the gradients to update the global top subnetwork and send it back to the active party. In this way, the regularized empirical risk minimization problem (1) can be rewritten as:

$$\min_{w_0, \boldsymbol{\theta}} F(w_0, \boldsymbol{\theta}),$$

$$F(w_0, \boldsymbol{\theta}) = \sum_{l=1}^{L} \frac{N_l}{N} F_l(w_0, \boldsymbol{\theta}),$$

$$F_l(w_0, \boldsymbol{\theta}) = \frac{1}{N_l} \sum_{n \in N_l} \ell(w_0, h_{n,1}, \ldots, h_{n,M}; y_n) + \lambda \sum_{m=1}^{M} r(\theta_m) \quad (2)$$

The problem (2) can be seen as the minimization problem of the sum of several vertical federated learning problems and each active party is aiming to minimize its own objective function $F_l(w_0, \boldsymbol{\theta})$ in vertical federated learning.

## 4.2 Proposed Objective Function

In real-world vertical federated learning tasks, different parties may have different computational and communication resources. We classify active parties into fast active parties and slow active parties according to their computational and communication resources. It will cause the straggler problem if active parties launch updates in a synchronous way. Therefore, it is natural to use asynchronous aggregation mechanism to reduce waiting time and make training efficient. Therefore, in CVFL, active parties launch updates asynchronously which indicates that both the global top subnetwork and all the bottom subnetworks are updated asynchronously. Although using asynchronous updating mechanism can alleviate the straggler problem, slow active parties still train slower than the fast active parties, which results in their labels contributing less to bottom subnetworks and the global top subnetwork and finally leads the trained models being biased to the fast active parties and performing badly. To an extent, this phenomenon is somewhat similar with the fairness problem in horizontal federated learning. That is, the slow active parties are treated unfairly due to the limited computational and communication resources. Comparing with the fast active parties, the gradients of slow active parties' objective function with respect to all bottom subnetworks and the top subnetwork take tiny weights in the aggregated models. The difference between our problem and the fairness problem in horizontal federated learning is that fairness problem in horizontal federated learning aims to improve the performance (e.g., accuracy for classification tasks) of the worst parties but our problem is mitigating the straggler problem and improving the performance of all parties.

To further address the straggler problem, we propose a novel objective function for slow active parties. Specially, we add a constant exponent $q$ that is greater than 1 to the slow active parties' objective functions, that is $F_l^q(w_0, \boldsymbol{\theta})$. By this way, the loss values of slow active parties will be increased and thus the gradients that slow active parties contributed to all bottom subnetworks and the global top subnetwork will also be amplified. In this way, the problem (2) can be rewritten as:

$$\min_{w_0, \boldsymbol{\theta}} F(w_0, \boldsymbol{\theta}),$$
$$F(w_0, \boldsymbol{\theta}) = \sum_{l \in L_{fast}} \frac{N_l}{N} F_l(w_0, \boldsymbol{\theta}) + \sum_{l \in L_{slow}} \frac{N_l}{N} F_l^q(w_0, \boldsymbol{\theta}) \quad (3)$$

Note that after adding a constant exponent $q$ ($q > 1$) to the slow active parties' objective functions, the gradients of the slow party's objective function with respect to the top subnetwork are $qF_l^{q-1}(w_0, \boldsymbol{\theta})\nabla_{w_0} F_l(w_0, \boldsymbol{\theta})$ and the gradients with respect to bottom subnetworks are $qF_l^{q-1}(w_0, \boldsymbol{\theta})\nabla_{\boldsymbol{\theta}} F_l(w_0, \boldsymbol{\theta})$. Comparing with objective function that without an exponent, the gradients are amplified by $qF_l^{q-1}(w_0, \boldsymbol{\theta})$ times. What's more, the coefficient $qF_l^{q-1}(w_0, \boldsymbol{\theta})$ can change dynamically along with the objective function $F_l(w_0, \boldsymbol{\theta})$. At the beginning of training, the objective function value is typically large and the gradients are amplified more correspondingly. As the training going on, the objective function value decreases and the gradients are amplified less. When the slow party's objective function value $F_l(w_0, \boldsymbol{\theta}) \leq q^{\frac{1}{1-q}}$, which indicates that the loss is small enough, in this case we set the $q = 1$ so that gradients of the slow active party are not be amplified.

---

**Algorithm 1:** CVFL for active party $l$ to actively launch dominated updates

**Input:** Local data $\{\{X_{n,l}\}_{n=1}^N, \{Y_{n,l}\}_{n \in \mathcal{G}_l}\}$ stored on $l$-th active party

1 Initialize the necessary parameters;
2 **while** *not convergent* **do**
3    **When** *receive $w_0$ from Collaborator* **then**
4       Update $w_0^l = w_0$
5    **end**
6    Pick an index $n$ from $\mathcal{G}_l$ randomly;
7    Collect $[h_1(\theta_1, x_{n,1}), h_2(\theta_2, x_{n,2}), ..., h_M(\theta_M, x_{n,M})]$ from all $M$ parties;
8    **if** *l-th party is slow active party* **then**
9       Collect $[r(\theta_1), r(\theta_2), ..., r(\theta_M)]$ from all $M$ parties;
10    **end**
11    Compute $\nabla_{h_m(\theta_m, x_{n,m})} F_l(w_0^l, \boldsymbol{\theta})$ for all $M$ parties;
12    Send index $n$ and $\nabla_{h_m(\theta_m, x_{n,m})} F_l(w_0^l, \boldsymbol{\theta})$ to all corresponding $M$ parties;
13    Compute $\nabla_{w_0^l} F_l(w_0^l, \boldsymbol{\theta})$;
14    Send $\nabla_{w_0^l} F_l(w_0^l, \boldsymbol{\theta})$ to collaborator;
15 **end**

---

## 4.3 CVFL Algorithm

As shown in Algorithm 1, active parties launch dominated updates asynchronously. For active party $l$, it picks up an index $n$ or a data mini-batch first and collect all corresponding embedding vectors $h_m(\theta_m, x_{n,m})$ from all $M$ parties. If the active party $l$ is slow active party, it also needs to collect the values of all regularization term $r(\theta_m)$ from all $M$ parties so as to calculate the gradients of its objective function with respect to all embedding vectors and its top subnetwork since we add an exponent $q$ to the slow parties' objective functions. Then the active party $l$ calculates the gradients of the objective function with respect to its local top subnetwork $\nabla_{w_0^l} F_l(w_0, \boldsymbol{\theta})$ and the gradients of the objective function with respect to all embedding vectors $\nabla_{h_m(\theta_m, x_{n,m})} F_l(w_0, \boldsymbol{\theta})$ for all $M$ parties. After calculating the gradients, the active party $l$ sends $\nabla_{w_0^l} F_l(w_0, \boldsymbol{\theta})$ to the collaborator and sends $\nabla_{h_m(\theta_m, x_{n,m})} F_l(w_0, \boldsymbol{\theta})$ together with index $n$ to the corresponding parties. All active parties will run Algorithm 1 asynchronously until training convergence.

As shown in Algorithm 2, when a party $m$ receives index $n$ and $\nabla_{h_m(\theta_m, x_{n,m})} F_l(w_0, \boldsymbol{\theta})$, it will launch a new collaborative update asynchronously. That is to say, it will calculate the gradients of its embedding vector with respect to its bottom subnetwork by chain rule as $\nabla_{\theta_m} h_m(\theta_m, x_{n,m}) \nabla_{h_m(\theta_m, x_{n,m})} F_l(w_0, \boldsymbol{\theta})$ and calculate $\nabla_{\theta_m} r(\theta_m)$. Then it updates its bottom subnetwork as shown in Step 5 in Algorithm 2. Note that active parties also need to perform Algorithm 2 to update their bottom subnetworks.

---

**Algorithm 2:** CVFL for party $m$ to passively launch collaborative updates asynchronously

**Input:** Local data $\{X_{n,m}\}_{n=1}^N$ stored on $m$-th party, learning rate $\gamma$

1 Initialize the necessary parameters;
2 **When** *receive an index $n$ and $\nabla_{h_m(\theta_m, x_{n,m})} F_l(w_0^l, \boldsymbol{\theta})$* **then**
3    Calculate $v = \nabla_{\theta_m} h_m(\theta_m, x_{n,m}) \nabla_{h_m(\theta_m, x_{n,m})} F_l(w_0^l, \boldsymbol{\theta})$;
4    Calculate $\nabla_{\theta_m} r(\theta_m)$;
5    Update $\theta_m = \theta_m - \gamma v - \gamma \nabla_{\theta_m} r(\theta_m)$;
6 **end**

---

As shown in Algorithm 3, when the collaborator receives $\nabla_{w_0^l} F_l(w_0, \boldsymbol{\theta})$, it will update the global top subnetwork asynchronously and send the updated global top subnetwork back to the corresponding active party $l$. Then when the active party $l$ receives the updated global top subnetwork, it will update its local top subnetwork $w_0^l$ using the received model as shown in the step 3-5 of Algorithm 1.

---

**Algorithm 3:** CVFL for the collaborator to update $w_0$ asynchronously

**Input:** learning rate $\gamma$

1 Initialize the necessary parameters;
2 **When** *receive $\nabla_{w_0^l} F_l(w_0^l, \boldsymbol{\theta})$ from active party $l$* **then**
3    Update $w_0 = w_0 - \gamma \nabla_{w_0} F_l(w_0^l, \boldsymbol{\theta})$;
4    Send $w_0$ to active party $l$;
5 **end**

---

## 5 EXPERIMENTS

In this section, we conducted comprehensive experiments to demonstrate the losslessness and efficiency of our proposed algorithm.

### 5.1 Experiment Settings

All experiments are conducted on a server that has 2 sockets and each socket has 12 cores. We use multithread to simulate the real environment that multiple parties involving in training neural networks. The features are partitioned vertically and equally into 4 non-overlapped parts since comparing with horizontal federated learning which has hundreds of thousands participants, vertical federated learning typically only has several participants. And recent vertical federated learning works [11] [24] that using image datasets to conduct experiments are usually setting there are no more than 4 parties that participate in model training. Therefore, we assume that there are total 4 parties that participating in model training in all experiments. The labels are also partitioned horizontally and equally into 4 parts. That is to say, the 4 parties are all active parties and they all need to train a bottom subnetwork and a local top subnetwork. The train data and test data are training dataset and testing dataset of the experiment data. We choose an optimal learning rate $\gamma$ from $\{1e^{-3}, 5e^{-3}, 1e^{-2}\}$ and we use the $L2$ regularization and set the coefficient $\lambda$ of the regularization term as $\lambda = 1e^{-3}$ in all experiments.

The datasets that we conducted experiments on are shown in Table 1. These three are all widely used image classification benchmark datasets. As illustrated before, the features are vertically partitioned into 4 parts, that is to say, the sizes of features that each party owns are $1 * 28 * 7$, $1 * 28 * 7$ and $3 * 32 * 8$ for the three datasets respectively. The labels are also partitioned equally into 4 parts and each party owns 15000, 15000 and 12500 labels for the three datasets respectively.

|  | MNIST | FashionMNIST | Cifar10 |
|---|---|---|---|
| #Training Samples | 60000 | 60000 | 50000 |
| #Testing Samples | 10000 | 10000 | 10000 |
| #Feature Dimensions | 1*28*28 | 1*28*28 | 3*32*32 |

Table 1: Dataset Descriptions

|  | MNIST | FashionMNIST | Cifar10 |
|---|---|---|---|
| Centralized Training | 99.01% | 92.67% | 82.51% |
| CVFL-single | 96.90% | 88.25% | 67.68% |
| **CVFL** | 98.28% | 91.51% | 77.90% |

Table 2: Accuracy of three algorithms to demonstrate the extent of losslessness and the effectiveness of our proposed algorithms. Each comparision is repeated 5 times to get the average.

### 5.2 Evaluation of Losslessness

To demonstrate the losslessness of our proposed CVFL algorithm in the scenario where labels are partitioned among active parties, we use CVFL, centralized training and CVFL-single to train models separately and compare the accuracy on the testing dataset. Centralized training is a training method that the features and labels are not partitioned and are located together in a machine.

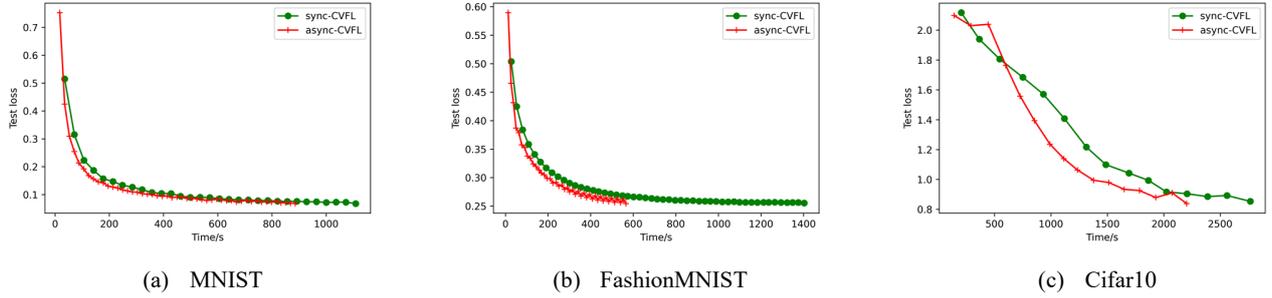

(a) MNIST  (b) FashionMNIST  (c) Cifar10

**Figure 2:** The test loss vs. wall-clock time curve to demonstrate the efficiency of asynchronous aggregation mechanism. Each point denotes an epoch's result.

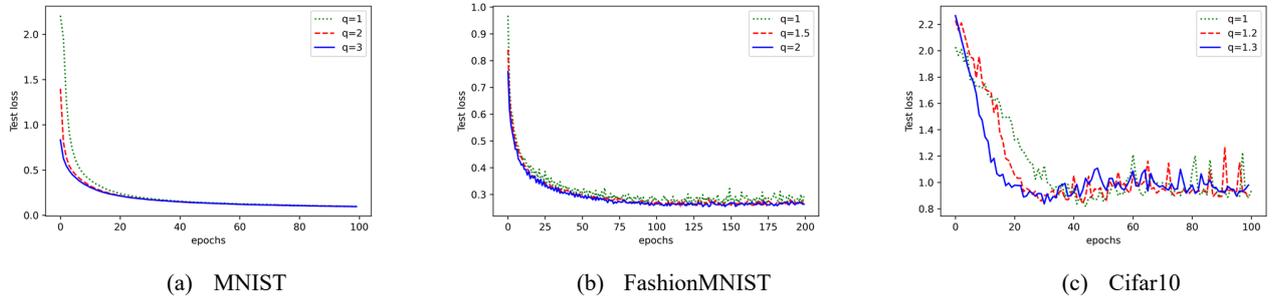

(a) MNIST  (b) FashionMNIST  (c) Cifar10

**Figure 3:** The test loss vs. epoch curve to demonstrate the efficiency of our proposed objective function, i.e., CVFL-$q > 1$ when distributions of labels are i.i.d. and slow active parties are 50% slower than the fast active party.

Typically, the performance of centralized training is the upper bound in federated learning. CVFL-single is the counterpart of CVFL but without the collaborator to perform top subnetwork aggregation. Therefore, CVFL-single can only use 1/4 number of labels to train models since it can only use a single part of labels in the scenario that labels are partitioned among 4 parties. The bottom subnetwork we used in three datasets is CNN so as to extract features from images and the top subnetwork is MLP. Note that both bottom subnetwork and top subnetwork can be any type of neural networks, since CVFL is a general framework and CVFL algorithm isn't coupled with the specific type of neural network. What's more, we keep the same bottom subnetwork and top subnetwork structures in CVFL, centralized training and CVFL-single so as to make sure the differences in experiment results are caused by the differences among training algorithms themselves. Each comparison is repeated 5 times and is stopped when training is converged. As shown in Table 2, the accuracies on test dataset of models trained by centralized training is the highest among the three methods and the accuracies of models trained by our proposed CVFL algorithm is very comparable with centralized training. Specially, on MNIST and FashionMNIST, the differences of accuracies are all less than 1.5% and on Cifar10, the difference of accuracy is less than 5%, which can demonstrate that CVFL is nearly lossless. Comparing with CVFL-single, CVFL outperforms 1.38% and 3.26% than CVFL-single on MNIST and FashionMNIST, respectively. Moreover, on Cifar10, CVFL outperforms over 10% than CVFL-single, which demonstrates the effectiveness of our proposed cascade training mechanism.

### 5.3 Evaluation of Asynchronous Efficiency

To demonstrate the efficiency of asynchronous aggregation mechanism, we compare it with synchronous counterpart of CVFL (denoted as sync-CVFL), i.e., active parties launch dominated updates synchronously. To simulate the real vertical federated learning tasks that computational and communication resources vary across active parties, we set that there are three active parties that are stragglers and are 50% slower than the fast active party.

In these experiments, we keep the model structure and learning rate the same in async-CVFL and sync-CVFL and we also set the same stop criterion, i.e., testing loss is less than $2.5e^{-1}$ for FashionMNIST dataset, for this two algorithms. As shown in Figure 2, the test loss vs. wall-clock time curves demonstrate the efficiency of async-CVFL since async-CVFL consistently outperforms sync-CVFL on three datasets. More specially, async-CVFL reaches the given stop criterion more than 20.1%, 57.1% and 20.4% earlier than sync-CVFL on these three datasets respectively.

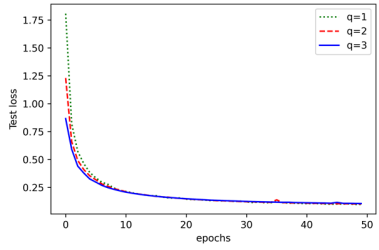 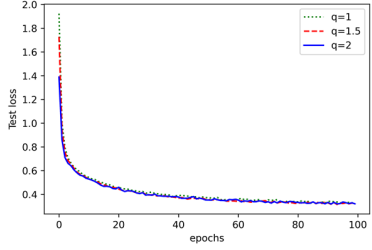 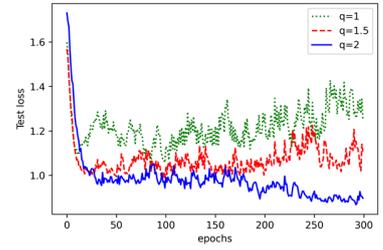

(a) MNIST  (b) FashionMNIST  (c) Cifar10

**Figure 4: The test loss vs. epoch curve to demonstrate the efficiency of our proposed objective function, i.e., CVFL-$q > 1$ when distributions of labels are i.i.d. and slow active parties are 95% slower than the fast active party.**

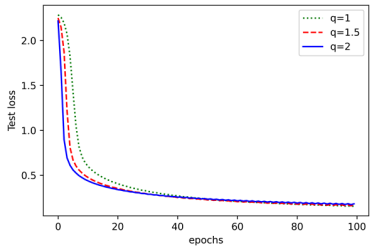 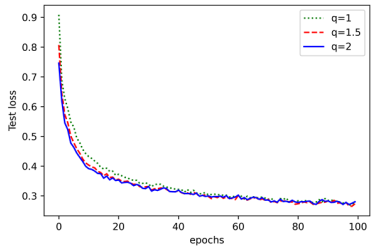 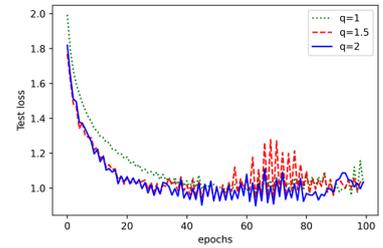

(a) MNIST  (b) FashionMNIST  (c) Cifar10

**Figure 5: The test loss vs. epoch curve to demonstrate the efficiency of our proposed objective function, i.e., CVFL-$q > 1$ when distributions of labels are non-i.i.d. and slow active parties are 50% slower than the fast active party.**

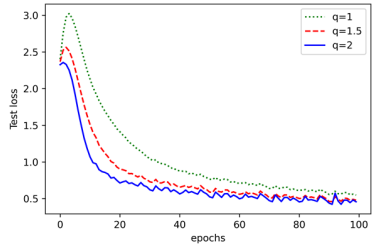 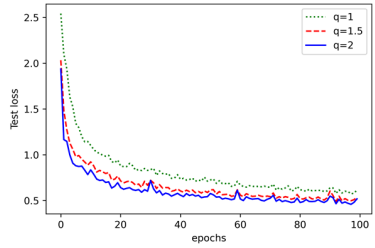 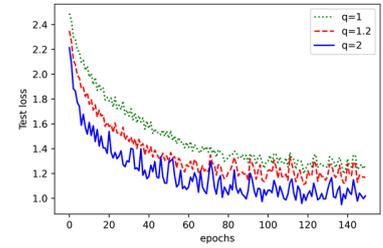

(a) MNIST  (b) FashionMNIST  (c) Cifar10

**Figure 6: The test loss vs. epoch curve to demonstrate the efficiency of our proposed objective function, i.e., CVFL-$q > 1$ when distributions of labels are non-i.i.d. and slow active parties are 95% slower than the fast active party.**

## 5.4 Evaluation of the Efficiency of the Proposed Objective Function

As illustrated before, only using asynchronous aggregation mechanism is not enough in the scenario where labels are partitioned among active parties. Therefore, we propose a novel objective function that with an exponent large than 1 (denoted as CVFL-$q > 1$) for slow active parties so as to further mitigate the straggler problem. To demonstrate the efficiency of our proposed objective function, we compare it with its counterpart that only using asynchronous aggregation mechanism but without an exponent large than 1 (denoted as CVFL-$q = 1$) added to slow active parties' objective function.

*5.4.1 Unbiased Distributions of Labels.* We first consider a simple situation that the distributions of labels are independent and identically distributed (i.e., i.i.d.) among active parties. As shown in Figure 3, when we set that there are three slow active parties and are 50% slower than the fast active party, CVFL-$q > 1$(e.g., $q = 3$) converges slightly faster than CVFL-$q = 1$ on MNIST and FashionMNIST datasets. Specially, CVFL-$q > 1$ converges obviously faster than CVFL-$q = 1$ on Cifar10 dataset. As shown in Figure 4, when we set that slow active parties are 95% slower than the fast active party, the results do not change greatly on MNIST and FashionMNIST datasets. The reason is that these two datasets are not overly complex and the distributions of labels are IID among active parties. Therefore, training converges very quickly on these two datasets. But on Cifar10 dataset, CVFL-$q > 1$ converges significantly better than CVFL-$q = 1$ since Cifar10 is a more complex dataset and when computational resources vary greatly across active parties, CVFL-$q > 1$ can amplify slow active parties' gradients to accelerate training convergence and make the trained model perform better (i.e., test loss is lower than CVFL-$q = 1$).

*5.4.2 Biased Distributions of Labels.* To simulate the real vertical federated learning tasks, we then conduct the following experiments in the situation that distributions of labels are non-i.i.d. among active parties. That is to say, different active parties own different classes of labels. More specially, for MNIST and FashionMNIST datasets, we set each party owns three different classes of labels and any two parties can only have at most one class of labels in common. For example, party 1 has labels of classes $\{0, 1, 2\}$ and party 2 has labels of classes $\{2, 3, 4\}$. For Cifar10 dataset, we set each party has 6 classes of labels in common, but each party owns a unique class of labels. For example, each party has labels of classes $\{0, 1, 2, 3, 4, 5\}$ in common, but party 1 has all labels of class 6 and party 2 has all labels of class 7. As shown in Figure 5, when we set that the three slow active parties are 50% slower than the fast active party, CVFL-$q > 1$ consistently outperforms CVFL-$q = 1$ on three datasets and we can see that the loss curves of CVFL-$q > 1$ drop faster than CVFL-$q = 1$, especially for the complex Cifar10 dataset. We then set that the slow active parties are 95% slower than the fast active party, the results are illustrated in Figure 6. We can observe that in this situation, the test loss curves of CVFL-$q > 1$ not only drop faster than CVFL-$q = 1$, but also achieve much lower test loss on these three datasets, which can strongly demonstrate the efficiency of our proposed objective function.

## 5.5 Discussion

In this part, we discuss the limitations of our work with possible solutions to overcome them.

First, although we have considered and addressed the straggler problem caused by slow active parties, the straggler problem caused by the slow passive parties are not considered in our work. Specially, when active parties launch model updates, they need to collect all embedding vectors from other parties and then take the concatenation of embedding vectors as input of their own top subnetworks. If there exists a slow passive party, the model updates process of the active parties will be substantially slowed down. A possible solution for the problem is allowing model update process to continue when certain number of embedding vectors have been collected and active parties can directly use the stale embedding vectors of the slow passive parties, instead of waiting for them.

Second, although our approach is not coupled with specific types of neural network or data, it does indeed face some challenges when dealing with different types of data, for example, sequential data. For sequential data, the partitioned features have sequence relations with each other. Therefore, the partitioned features that are in front of the sequence must be learned before the partitioned features that are behind them. In this case, parties have to collaborate with each other in a sequential way and active parties cannot launch model updates asynchronously, which will make training process inefficient. There still needs further work to handle this problem.

## 6 CONCLUSION AND FUTUREWORK

In this paper, we have developed a cascade vertical federated learning framework (CVFL) that enables to train neural network in the scenario where features and labels are all partitioned among different parties. CVFL breaks down the entire neural network into two types of subnetworks so that it can utilize the partitioned labels to train neural network in a privacy-preserving way, which is not supported by existing vertical federated learning methods. Moreover, we proposed a novel objective function which is with an exponent large than 1 so as to further alleviate the straggler problem. Through the experiments on three widely-used image classification datasets, we have demonstrated the losslessness and efficiency of our proposed algorithm. In future work, we plan to conduct more experiments that using different types of neural networks, for example, using RNN or LSTM to extract features from text data and using Graph Neural Network (GNN) to extract features from graph data. Another research direction is to design a proper algorithm that can automatically select the value of exponent $q$ and dynamically change it as the training goes on.